\documentclass[final,5p,times]{elsarticle}
\usepackage{amssymb}
\usepackage{graphicx}
\usepackage{float}
\usepackage[T1]{fontenc}
\usepackage{amsmath,color}
\usepackage{hyperref}
\journal{Computers \& Graphics}

\begin{document}

\begin{frontmatter}

\title{\textbf{Fast color transfer from multiple images}}

\author{Asad Khan, Luo Jiang, Wei Li, Ligang Liu}
\address{Computer Graphics Laboratory, School of Mathematical Sciences,\\
University of Science and Technology of China,\\
Hefei, Anhui, 230026, PR China}
\begin{abstract}
Color transfer between images uses the statistics information of image effectively.
We present a novel approach of local color transfer between images based on the
simple statistics and locally linear embedding. A sketching interface is proposed
for quickly and easily specifying the color correspondences between target and
source image. The user can specify the correspondences of local region using scribes,
which more accurately transfers the target color to the source image
while smoothly preserving the boundaries, and exhibits more natural
output results. Our algorithm is not restricted to one-to-one
image color transfer and can make use of more than one target images to transfer the
color in different regions in the source image.
Moreover, our algorithm does not require to choose the same color style and image size between source
and target images. We propose
the sub-sampling to reduce the computational load. Comparing with other approaches,
our algorithm is much better in color blending in the input data.
Our approach preserves the other color details in the source image. Various experimental results show that our
approach specifies the correspondences of local color region in source and target images.
And it expresses the intention of users and generates more actual and natural results of visual effect.
\end{abstract}

\begin{keyword}
Image processing\sep Local color transfer\sep Locally linear embedding\sep Edit Propagation
\MSC I.4.9 [Image Processing and Computer Vision]

\end{keyword}

\end{frontmatter}

\section{Introduction}

Color transfer is an image processing method imparting the color characteristics
of a target image to a source image. Ideally, the result by a color transfer
algorithm should apply the color style of the target image to the source image.
A good color transfer algorithm should provide quality both in scene details and colors.\\

Reinhard et al. \cite{1} presented a simple and potent color
transfer algorithm which translates and scales an image
pixel by pixel in $L\alpha\beta$ color space according to the mean
and standard deviation of the color values in the source and target images.
There exist numerous procedures in literature where probability distributions
are used to process the image's colors \cite{3,15,20} or deliver
user controllable adjustment of the colors. Out of these, the latter
ones are either restricted to local editing \cite{42} or contain global
edit propagation \cite{41}. These procedures have been proven to provide
most satisfactory results, but their common drawback is usually a
somewhat large computational load because of global optimization.\\

The technique we develop in this paper is local color transfer between
images utilizing the statistical information of image effectively.
We present a new method of local color transfer based on the simple
statistics and locally linear embedding (LLE) to optimize the constraints in a
newly defined objective function of images. Our procedure will automatically
\begin{figure*}[t!]
\centering
  \includegraphics[width=1.0\textwidth]{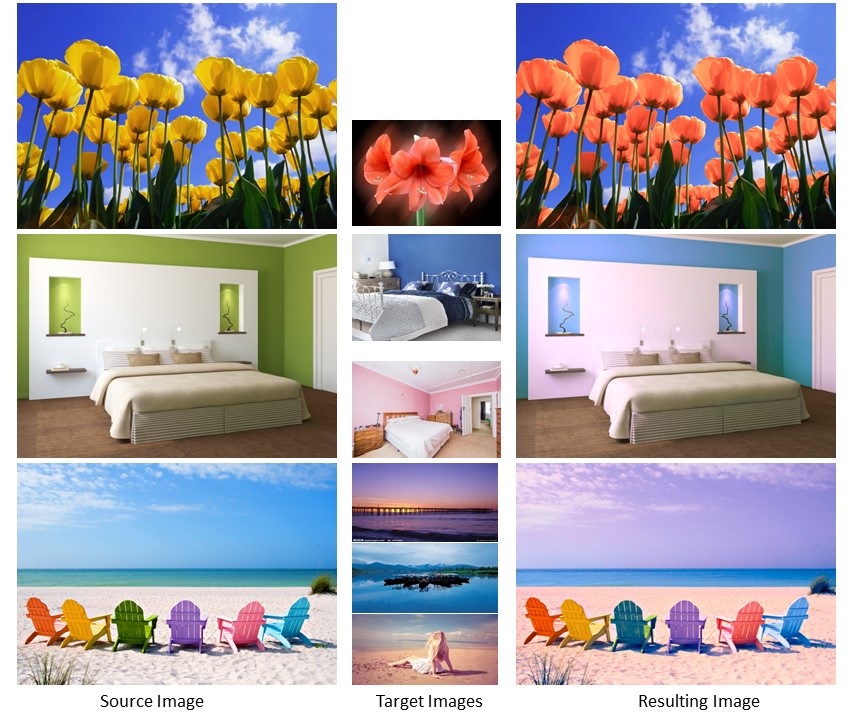}\\
  \caption{Application of three different environments and boundary preservation.}
  \label{fig:figure1}
\end{figure*}
determine the influence of edited samples across the whole image
jointly considering spatial distance, sample location and appearance. We will convey
local color of an image to others using LLE.\\

Although, in previous methodologies,
rough strokes followed by LLE were used for propagation of the color \cite{37}. Our procedure
differs from others in the sense that we formulate a new methodology for the
optimization problem where we draw from the non-linear manifold learning
formulation. We excogitate the problem as a global optimization task and
show that it can be solved as a sparse linear system. This merges global
editing as in An and Pellacini \cite{41}, who used a dense solver, and a sparse
optimization borrowed from Lischinski et al. \cite{42}. We then modified it to our needs.\\

The inspiration of their work was also from the manifold-learning methodology \cite{39},
however, An and Pellacini \cite{41} exhibited that this work was not suitable
for propagating high quality images. Instead, An and Pellacini gave a methodology
which uses a dense least-squares solver, that allowed them to propagate the
affinities of all pairs of pixels to each other in order to retain the quality.
However, their dense linear system most often does not fit into the computer
memory for usual images. On the other hand, the method of Lischinski et al. \cite{42}
employs a sparse solver to deliver high-quality image propagation, but it applies
the edits only to nearby pixels which are spatially coherent.
It also needs a better accuracy in user input to achieve good results.\\

 In this paper we introduce a formulation of the optimization which tries
 to achieve both a global pixel interaction along with sparse solution.
 To accomplish this we interpret the image color as a manifold in 3D space
 by using the locally linear embedding algorithm \cite{40}. Then our work builds
 on that and automatically determines the influence of edit samples across
 the whole image jointly considering spatial distance, sample location and
 appearance. We show how a color manifold can be warped globally to obtain
 recoloring while its local relationships are conserved in order to retain
 the appearance of the source image.\\

In addition, we introduce another
stratagem in order to achi-\\eve interactive performance. We sub-sample
the image which greatly reduces the number of color points to be considered.
Next, we approximate the manifold using the sub-sampled points and interpolate
the remaining values. Then we maintain the same sub-sampling for different
user inputs and only update the target color values which are provided by
the user. Our procedure has a small memory usage and scales linearly in
the number of pixels while allowing interactive editing.\\

In short summary, this article makes the following contributions:\\
$\bullet$ local color transfer between images based on the simple
statistics and locally linear embedding (LLE), which more accurately transfers
the target color to the source image while preserving
the boundaries and exhibits more natural output results,\\
$\bullet$ our algorithm is not restricted to one-to-one image and can
have more than one target images to transfer the color in different regions in the source
image, and\\
$\bullet$ we also propose the sub-sampling which reduce in order to computational load.\\


\section{Related Work}

Now a days, color transfer is a much debated research area.
We can classify these color transfer techniques in two algorithms,
namely global and local algorithms.\\

Reinhard et al. \cite{1} and his colleagues were the first to implement a color
transfer method by globally transferring colors, after translating color
data of input images from the RGB color space to the decorrelated $L\alpha\beta$
color space. This transferred colors quickly, successfully and also efficiently
generated a convincing output. This technique was further improved by Xiao et al. \cite{14}.
Pitie et al. \cite{15} who used a refined probabilistic model.
In Pitie et al. \cite{4} they furthered their method in order to better
perform non-linear adjustments of color probability distribution between images.
Similarly, Chang et al. \cite{16,17} suggested global color transfer by introducing
perceptual color categorization for both images and video. Yang et al. \cite{18}
initiated a method for color mood transfer which preserves spatial coherence
based on histogram matching. This idea was developed further by Xiao et al. \cite{19}
who puzzled out the problem of local fidelity and global transfer in two steps: gradient preserving optimization
and histogram matching. Wang et al. \cite{20,21} proposed a technique for global
color-mood exchange driven by predefined and labeled color palettes and example images.
Cohen  et al. \cite{22} suggested a methodology which employs color harmony
rules to optimize the overall appearance after some of the colors have been
altered by the user. Shapira et al. \cite{23} suggested a solution which utilized
navigating the appearance of the image to obtain desired results. Furthermore,
automatic methods for colorizing grayscale images stemming from examples from internet
images \cite{24} and semantic annotations \cite{25} were introduced.
In general, color transfer methods which act globally are not competent
enough for accurate re-coloring of small objects or humans.\\

Other approaches tried to counter above mentioned shortcomings by
introducing rough control in image editing. Various distance measures
and feature spaces were also considered in the literature. To cross
texture and fragment distinct regions, geodesic distance \cite{28} and
diffusion distance \cite{29} were applied. Li et al. \cite{30}
championed the use of pixel classification based on user input.
Locally linear embedding propagation preserved the manifold structure \cite{37,38} to tackle color blending.\\


To prevent the problem relating to color region mixing, Neumann et al. \cite{2}
suggested a 3D histogram matching technique to transfer
color components in the hue (H), saturation (S), and intensity (I), respectively,
in the HSI color space. Albeit, the color information transfer between target
and source image can be achieved by this method, the result usually contains
notable spatial artifacts and is also dependent on the resolution of input image.
Pitie et al. \cite{3,4} resolved the same problem by utilizing an N-dimensional
probability density function which matches the 3D color histogram of input images.
They used an recursive, nonlinear method that was able to estimate the transformation
function by employing a one-dimensional marginal distribution. This technique is
potent enough to match the color pallet of target and source images, but it often
demands further processing to get rid of noise and spatial visual artifacts.\\

Color region mixing problem resolution and transference of colors to a local region
of an image are problems for which the segmentation-based techniques have been developed.
Tai et al. \cite{5,6} attempted to provide a solution to these problems by
employing a method for soft color segmentation which is based on a mixture of
Gaussian approximation and allows indirect user control. 
To resolve the color region mixing problem, some researchers tried to segment
input images by employing a fixed number of color classes  \cite{7,8} which uses a similar approach by Tai et al. \cite{5,6}.
But when the color styles of input images differ considerably
from the reference color classes, these approaches lack
somewhat in providing appropriate segmentation results. Yoo et al. \cite{27}
also used soft segmentation for local color transfer between images.
They have tried to solve color region mixing problem, but one drawback of
their method is that while transferring the color to a local region, the color of other regions also get effected.
Now, we intend to solve this problem by using our
algorithm to transfer the color from the target image to the source
image without effecting the colors in other regions.\\

For transferring colors among desired regions only, manual
approaches with user interventions were suggested by some researches.
Maslenicova et al. \cite{9} defined a rectangular area in each input
image where color transfer was desired, then utilizing region propagation,
they generated the color influence map. Pellacini et al. \cite{10} suggested a stroke-based
color-transfer technique which employs pairs of strokes to specify
the corresponding regions of target and source images. Although it is
possible for users to change the color of a local region by using some
strokes, it may call for strenuous efforts for detailed doctoring such
as oil paintings and complex images.\\

Recently, Baoyuang et al. \cite{11}
proposed color theme enhancement of an image. They effectuated a new
color style image by using predefined color templates instead of source
images. To perform decently, it needs quite accurate user input.
The color transfer methodology is also utilized to apply colors to grayscale images.
Tomihisa et al. \cite{12} assigned chromaticity values by equating the
luminance channels of target and source images. Abadpour et al. \cite{13}
yielded reliable results by employing a principal component analysis method.\\

\begin{figure*}[t!]
\vspace{-0.1cm}
\centering
  \includegraphics[width=18cm,height=9cm]{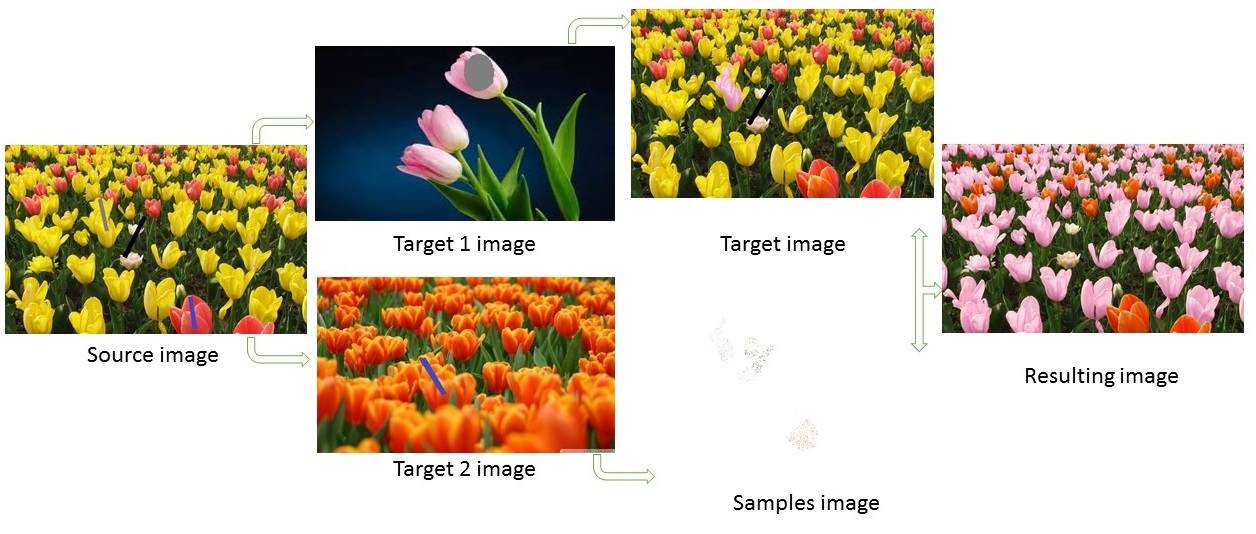}\\
  \caption{The pipeline of our algorithm-refer to Sect. 3 for details. From
left to right: source image in the first column and target images in the second column. The pink color
in Target 1 image is intended to transfer on the yellow color in the source image, where the gray spots on
both these colors shows this correspondence. Similarly, the dark orange color in the Target 2 image is desired to transfer on the light red
color in the source image, while the blue spots show this map. Whereas, the colors under black spot in the source image are desired to remain
unchanged. Column three shows the Target image after the color transfer in first row, and Sample image in the second row. The final result after sub-sampling and propagation of colors is depicted in the final column.}
\label{fig:figure3}
\end{figure*}
Moreover, some researchers have exhibited a keen interest in transformation
of colors among distinct color spaces. Color transfer technique
warrants the use of a color space where major elements of a color are
mutually independent. Since, in the RGB color space, the colors are correlated ,
the decorrelated CIELab color space is usually employed.
This requires a method to effectuate color transfer transformation of the
color space, RGB to CIELab and vice versa. Xiao et al. \cite{14} proposed an improved solution
that circumvents the transformation process between the correlated color
spaces and uses translation, rotation, and scaling to transfer
colors of a target image in the RGB color space.\\

The method of Chen et al. \cite{37} is
based on the local linear embedding \cite{39,40} so is our approach. The main difference of our algorithm to their's is the
difference in methodology of local color transfer between images based on the simple
statistics and locally linear embedding (LLE) to optimize the constraints in a
defined objective function of images. Our algorithm that preserves
pairwise distances between pixels in the original image.
It also simultaneously maps the color to the user defined target
values and then use the sub-sampling to reduce the computational load. \\

\section{Formulations for Local Color Propagation}



\noindent
\textbf{3.1   Local color transfer}\\

Give a source and target image, we can transfer the color from region $R_t$ in target image to region $R_s$ in source image by the following equation
\begin{equation}\label{local_color_transfer}
  {s_i^c}^{\ast} = \frac{\sigma_t^c}{\sigma_s^c}(s_i^c - \mu_s^c) + \mu_t^c, \quad c = l, \alpha, \beta, i\in R_s
\end{equation}
where $s_i^c$ and ${s_i^c}^{\ast}$ are the initial and final value of source image in channel $c$. And $\mu_s^c$ and $\mu_t^c$ are the averages of the values of channel $c$ in $L\alpha\beta$ color space for source and target image, respectively, and $\sigma_s^c$ and $\sigma_t^c$ are the standard deviations of the values of channel $c$ for source and target image, respectively. And $R_s$ is the mask region in source image.\\

\noindent
\textbf{3.2   Locally Linear Embedding}\\

Our algorithm is inspired by the Locally Linear Embedding
(LLE) \cite{39}, that eliminates the need to estimate pairwise
distances between widely separated data points. LLE enterprises from a high dimensionality
to a lower dimensionality manifold settled on the simple
intuitional that every sample can be interpreted by a linear combination
of its neighbors. Let us suppose a vector $x_i$ to represent a
pixel $i$ in some feature spaces. Given a data set $x_1, ... , x_N,$ for each
$x_i$, we find its $k$ nearest neighbors, namely $x_{i1}, ... ,x_{ik}$. We compute
a set of weights $\omega_{ij}$ that can best rebuild $x_i$ from these $k$
neighbors. LLE computes $\omega_{ij}$ by minimizing
\begin{figure*}
\centering
  \includegraphics[width=18cm,height=22cm]{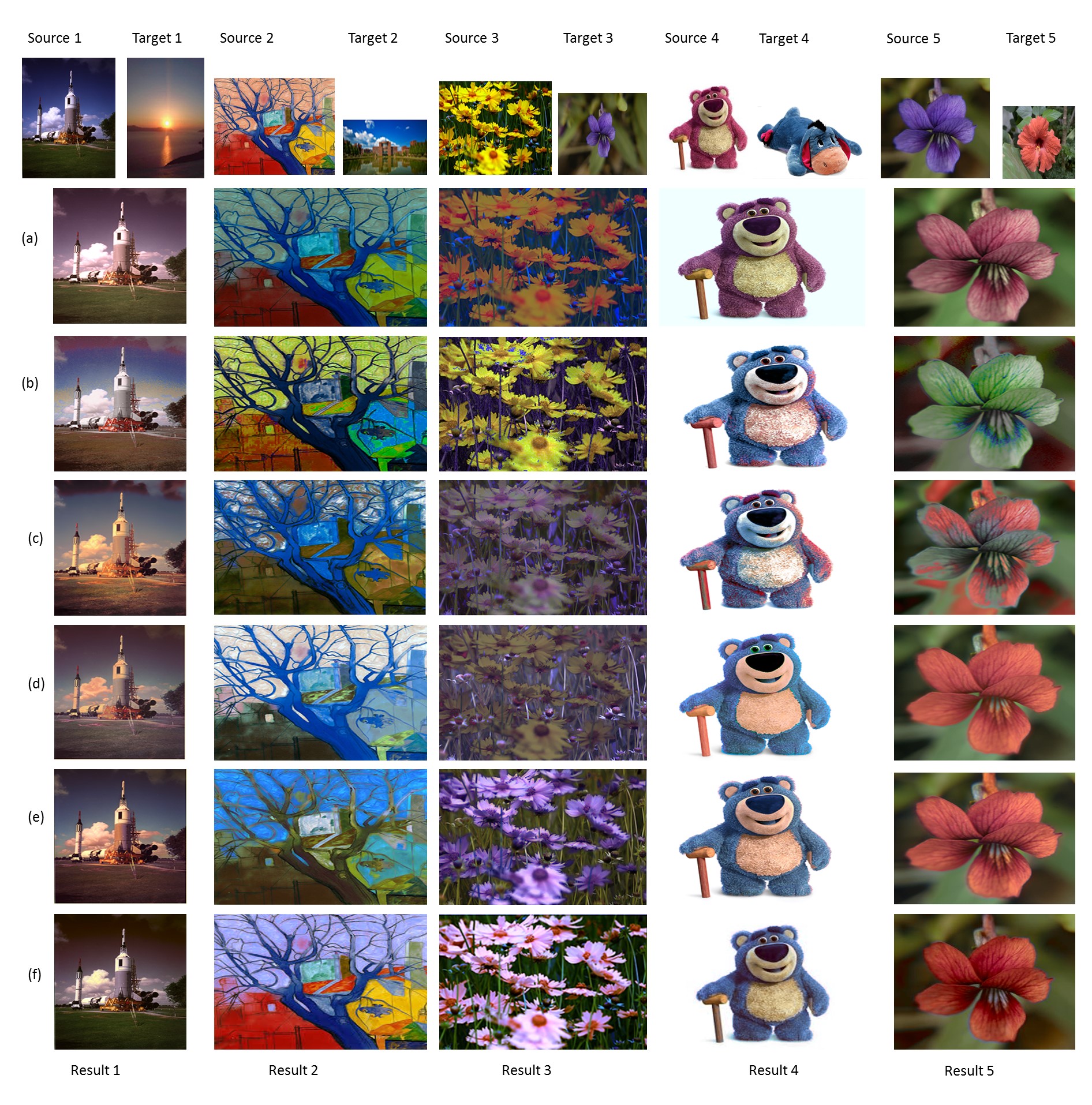}
  \caption{Comparing results with technique purposed before: (a) [Reinhard et al. 2001], (b) [Neumann et al. 2005], (c) [Pitie et al. 2007], (d) [Tai et al. 2007], (e) [Yoo J.-D. et al. 2013] (f) our implemented method.}
  \label{fig:figure6}
\end{figure*}
\begin{figure*}
\centering
  \includegraphics[width=16cm,height=6cm]{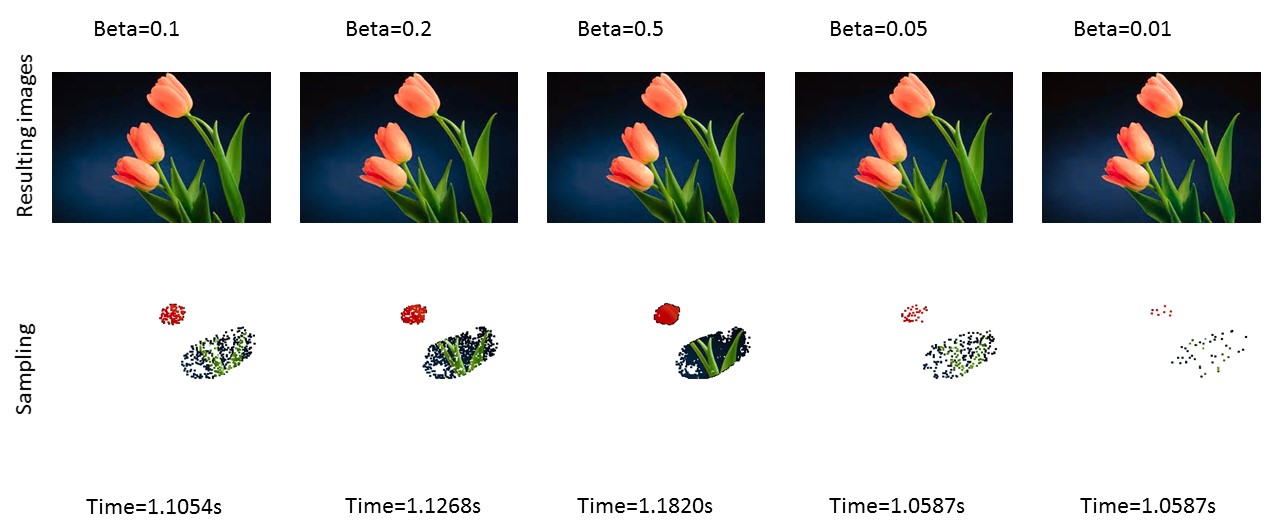}\vspace{-0.1cm}\\
  \caption{Comparison of the influence of the parameter $\beta$ on the results corresponding to a target image. We performed experiments with different values of $\beta$ and examined its optimal value where it gives both the better speed and the better quality. The second row shows
the actual landmarks sampled with respect to $\beta$ by taking $k=30$ in this example.}
\label{fig:figure11}
\end{figure*}

\begin{equation}\label{LLE}
  \sum_{i=1}^N || x_i-\sum_{j \in N_i} \omega_{ij}x_{ij} ||^2,
\end{equation}
subject to the constraint $\sum_{j \in N_i} \omega_{ij} = 1$. Then from the weights we can reconstruct $x_i$ from its neighbors.\\

\noindent
\textbf{3.3   Color propagation}\\

Given a source and target image, we can propagate color by minimizing the following energy
\begin{equation}\label{color_propagation}
  E = \sum_{i=1}^N(s_i-\sum_{s_j \in N_i}\omega_{ij}s_j)^2 + \lambda \sum_{i \in R}(s_i-t_i)^2
\end{equation}
where $s_i$ and $t_i$ are the value of pixel $i$ in source and target images, respectively. And $N_i$ is the neighborhood set of pixel $i$. $R$ is the region in target image whose color needs to propagate in source image. And $\lambda$ is a parameter that determines the relative importance of the second term compared with the first term.
\\
This energy can be further written in a matrix form as
\begin{equation}\label{color_propagation_matrix}
  E = [(I-W)S]^T(I-W)S + (S-T)^T \Lambda (S-T)
\end{equation}
and here $S$ is a vector with the $i_{\mathrm{th}}$ element $s_i$, $I$ is the identify matrix. And $\Lambda$ is a diagonal matrix with the $i_{\mathrm{th}}$ diagonal element $\lambda$ if $i \in R$. $T$ is a vector with the $i_{\mathrm{th}}$ element $t_i$ if $i \in R$. So the minimization of the energy is equivalent to solving a sparse linear system a follows.
\begin{equation}\label{color_propagation_linear}
    [(I-W)^T(I-W) + \Lambda ] S = \Lambda T
\end{equation}

\textbf{3.4   Sub-Sampling}\\

We design the algorithm in a way that it is suitable to work with all
pixels in the image according to the assumed facts.
Unluckily, the algorithm would be requiring the target values for
all pixels which is so tiresome and not desirable to provide such targets.
Moreover, it also requires a very high computational time. To deal with
a significant reduction of computational load and sparse target values,
our strategy is a sub-sampling approach which deal with both of the
above discussed problems. The expression of all the color points
by the linear combination of other points is the observation by making it
a base.\\

Therefore, the idea is to compute a number of substantial
sample points/landmark points and applying optimization
merely on those points. Then, the linear combination of the
landmark points is used to reconstruct the remaining points.
Depending on the application, different sampling strategies may be considered.
But so far random sampling \cite{31} has been the standard.
Random sampling may work well when the sample size is sufficiently large. But
large sample size will increase the workload.
In the applications random sampling while still achieving good performance.\\

We determine the landmarks using the original point set
$s$: we draw a random index set $\eta$ 
from the full index set $\tau = \{1 . . . N\}$ of all points. In order to
get significant points into $\eta$ , we require the chosen points
$s_j$ to be $(1)$ unique and $(2)$ linearly independent such that
they form a (generalized) Delaunay triangulation in $R^D$. For
each of the remaining points $s_i$ in the set $\{i|i\in \tau \backslash\eta \}$ we
determine the (D +1)-dimensional simplex $S$ in which it is
contained and compute its linear coefficients $L_i$ with respect
to $S$. Now, all points $s_i$ can be reconstructed as linear combinations
of the vertices of their Delaunay-simplices, thus,
$L_i$ are in fact barycentric coordinates. Note that they have to
be computed only once in the preprocessing stage.\\

Now we solve the problem of Eq. \ref{color_propagation} only for the landmark
points $\{t_j |j \in \eta\}$ and all other points $\{t_i |i \in \tau\backslash\eta \}$
are computed as linear combinations of the known points $t_j$
using the previously computed linear coefficients $L_i$. Also
the target values can be assigned in a user interaction pass to
landmarks points $\{t_j |j \in¸\eta \}$ only.\\

We control the ratio $\beta$ by the sub-sampling rate of the
points which directly affect the computational speed and
more importantly the quality of of the output images.
The principle is that to get the better estimation of underlying manifold,
requires more landmark points have to be used.
The longer computational time is the main flaw then.
It has been observed in the experiments we have performed so far
that the value of $\beta=0.05$ is better tradeoff between quality and speed. Fig. \ref{fig:figure11} depicts this relationship.
\begin{figure*}[t!]
\centering
  \includegraphics[width=18cm,height=9cm]{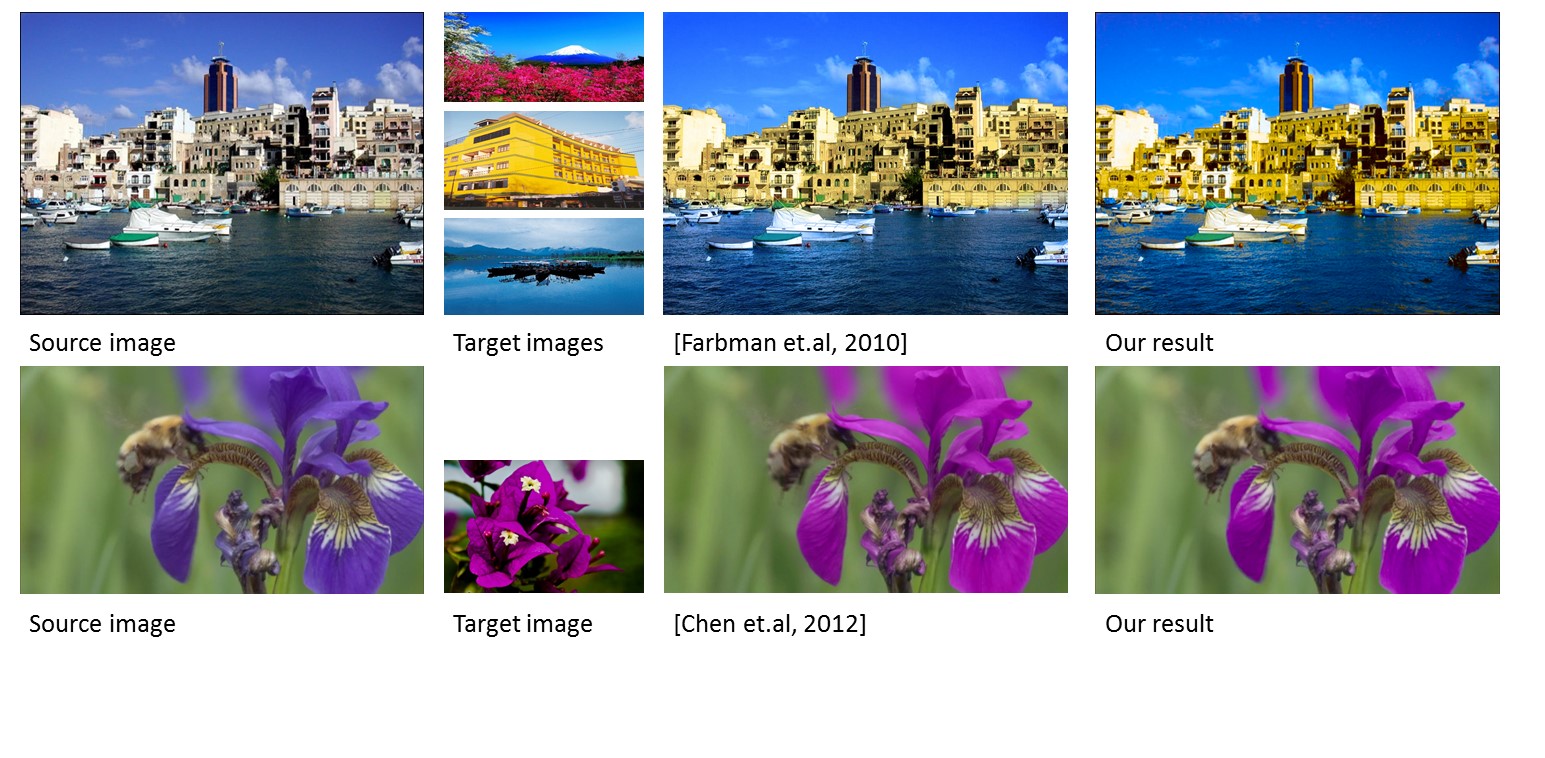}\vspace{-1cm}\\
  \caption{Comparison with the results of [Farbman et al. 2010] and [Chen et al. 2012].}
  \label{fig:figure8}
\end{figure*}

\section{Results and applications}
The experiment was performed in the computing environment
with Matlab2014a on a PC with an Intel(R) Core (TM)
i5-4690 CPU, 3.50 GHz processor and 8GB RAM under Windows OS.
Furthermore, the time taken by our algorithm with source image of $1024\times686$ pixels and two target images
of $1024\times768$ and $301\times220$ pixels by setting $k = 21$ is about 14.18 seconds.
We set $k=21$ in all the experiments shown in this paper. Our system uses
freehand closed regions to select and transfer colors from each target images to
source image. Then we select some regions in the source image
where the color needs to be transferred and
some regions where it does not required be changed.
This selection of regions can easily be drawn with the mouse freehandedly.
The user interface is quite easy to use even for novice users. Fig. \ref{fig:figure3} depicts this relationship.\\

The experimental results of our proposed method are compared
with those of the existing methods, such as the methods
of Reinhard et al. \cite{1}, Neumann et al. \cite{2}, Pitie et al. \cite{4}, Tai
et al. \cite{6} and Yoo et al \cite{27}.
The comparison is also made with the results of strokes based techniques
used by Farbman et al. and Chen et al. in \cite{29} and \cite{37} respectively.\\

Our suggested technique with the previous existing techniques were
tested using six different pairs of images that include landscapes
and objects as shown in Fig. \ref{fig:figure6}. The first row of the Fig. \ref{fig:figure6} indicates the
source image, on the left, and the target image, on the right.
The results of each technique are shown for Reinhard et al. \cite{1} in (a),
Neumann et al. \cite{2} in (b), Pitie et al. \cite{4} in (c), Tai et al. \cite{6} in (d),
Yoo et al. \cite{27} in (e) and our implemented results in (f) respectively.\\

The comparison with the results of Reinhard et al. is made in Fig. \ref{fig:figure6}(a). They transfer the local
color between images, but they were not able to control the transference of color in some regions
where the color needs not to be transferred which results in producing some artifacts as shown in Fig. \ref{fig:figure6}(a).\\

\begin{figure}[h!]
\centering
  \includegraphics[width=9cm,height=9cm]{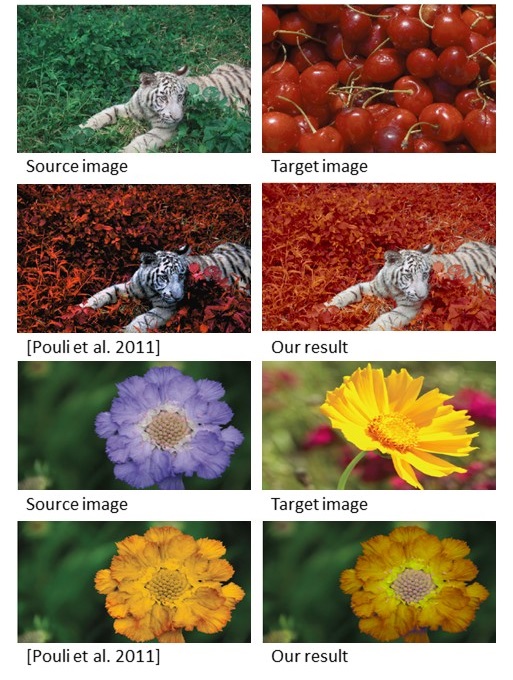}\vspace{-0.1cm}\\
  \caption{Comparison with the results by [Pouli et al. 2011].}
  \label{fig:figure9}
\end{figure}

The results of Neumann et al. and Pitie et al., are shown in Fig. \ref{fig:figure6}(b) and \ref{fig:figure6}(c) respectively. 
Both of these are actually the histogram-based
local color transfer methods. The drawbacks of their methods are including the transference of color in
unnecessary regions and the unexpected change in color style after its transference.
This is because of the color mapping which is luminance-based and different distribution of colors, results into
a blur and noisy images. The other reason seems to be the color mapping which is carried out
within pixels of similar luminance value.\\

The last two results by Tai et al. and Yoo et al., which are segmentation-based methods,
are depicted in Fig. \ref{fig:figure6}(d) and \ref{fig:figure6}(e). Their method matches regions of similar luminance
value hence the tree of the oil painting image in Result 2 is matched to the sky
region of the target image. The Result 3 shows that the
flowers have different luminance values and regions compared
to the input images as the intuitive region matching
between flowers was not performed properly.\\

The resulting images of our proposed method are depicted in Fig. \ref{fig:figure6}(f).
The comparison between our method and the other existing methods are shown in Fig. \ref{fig:figure6}.
Our proposed method transfers the target color to the source image while
preserving the boundaries, and exhibits more natural output results.
Intuitively, the starting region matching is made between meaningful regions
regardless of the difference in colors and luminance distributions
as shown in Fig. \ref{fig:figure6} by the region of sky in Result 1, the oil painting image in Result 2,
flowers in Result 3 and Result 5, and the toys in Result 4. The restriction of the one-to-one region matching
is not required by our method, since the number of dominant input images are not always
the same. Our method also excludes minor colors since the
one-to-one matching does not guarantee a satisfactory result
when the color styles of source images are quite different.\\

\begin{figure*}[t!]
\centering
  \includegraphics[width=1.0\textwidth, height=0.95\textheight]{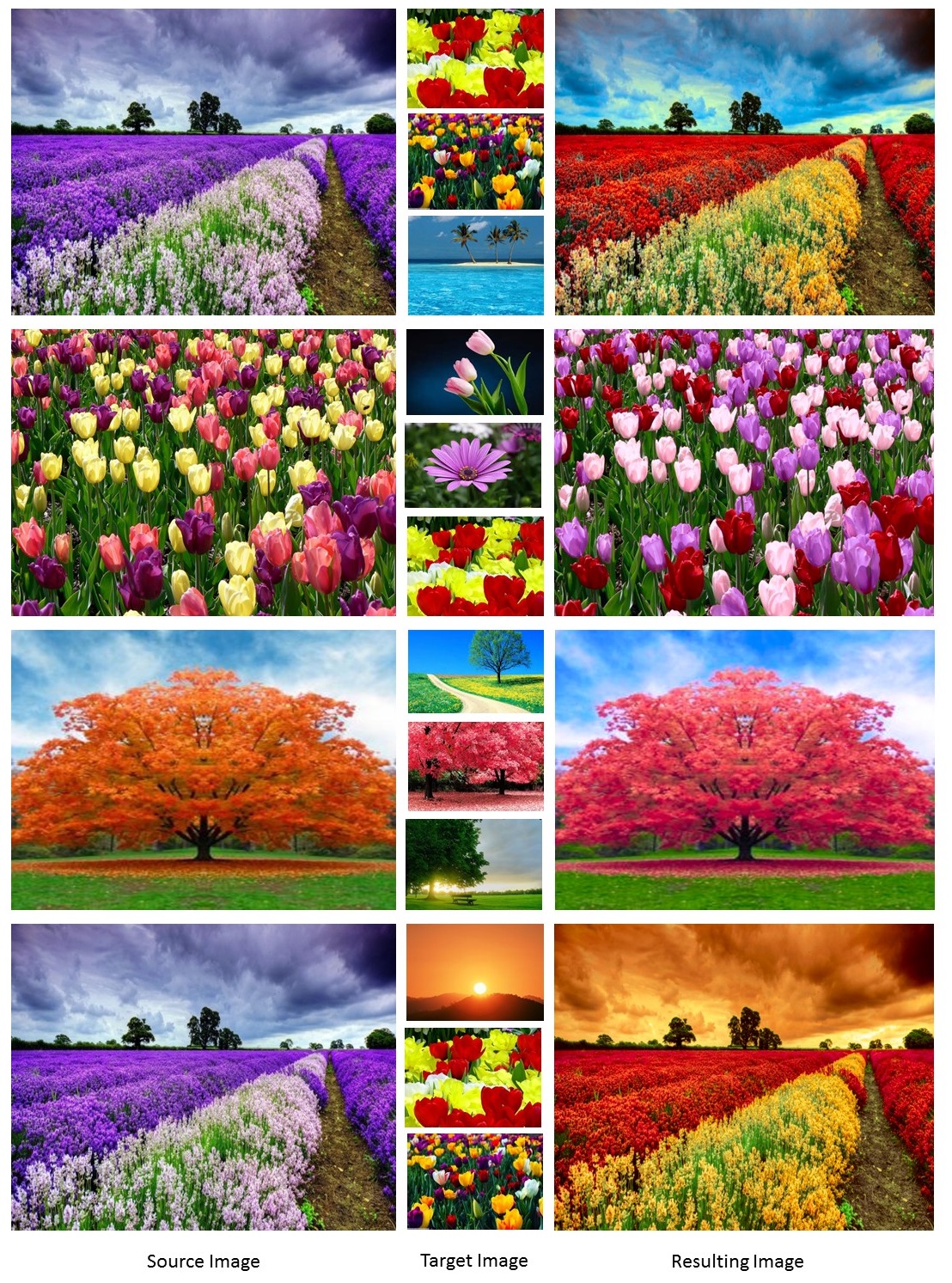}\\
  \caption{Multi-pronged application of our method.}
  \label{fig:figure7}
\end{figure*}

In our method, the focus is being put to preserve the boundaries
in the resulting image and to control the color expansion to the regions
where it is not required to be transferred. As a result, the quality of image remains
better as it can be observed from the comparison results in Fig. \ref{fig:figure6}. The color expansion
to other unnecessary regions makes the image blur with the compromise in its quality.
Our proposed method focuses to solve this problem as it is clear from our results.
Our algorithm is not restricted to one-to-one image and can have more than one target images to
transfer the color in different regions in the source image. Consequently, it provides more efficient,
natural and convincing results. The results in the different environments with more than one target images
are depicted in Fig. \ref{fig:figure1} and \ref{fig:figure7}.\\

 \begin{figure*}[t!]
\centering
  \includegraphics[width=1.0\textwidth, height=0.95\textheight]{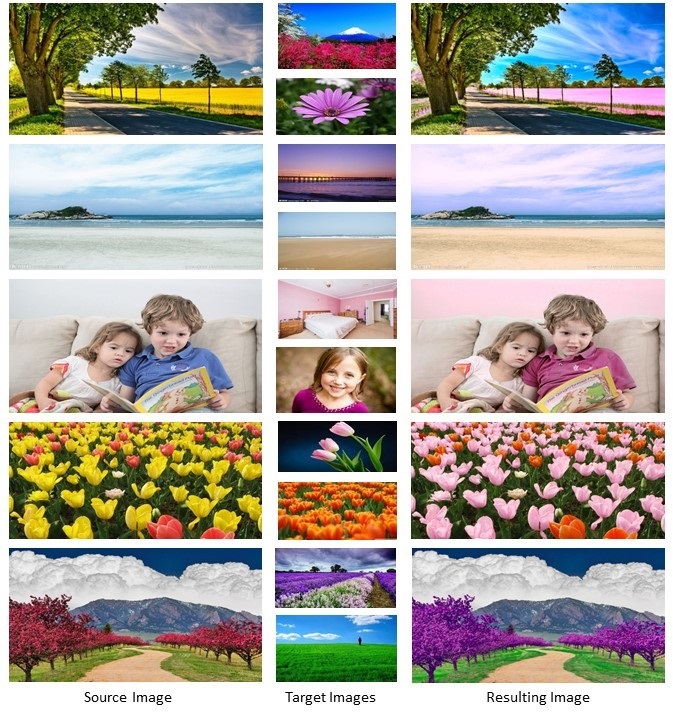}\\
  \caption{Results of one source image and two target images.}
  \label{fig:figure4}
\end{figure*}
\begin{figure*}[t!]
\centering
  \includegraphics[width=1.0\textwidth,height=0.95\textheight]{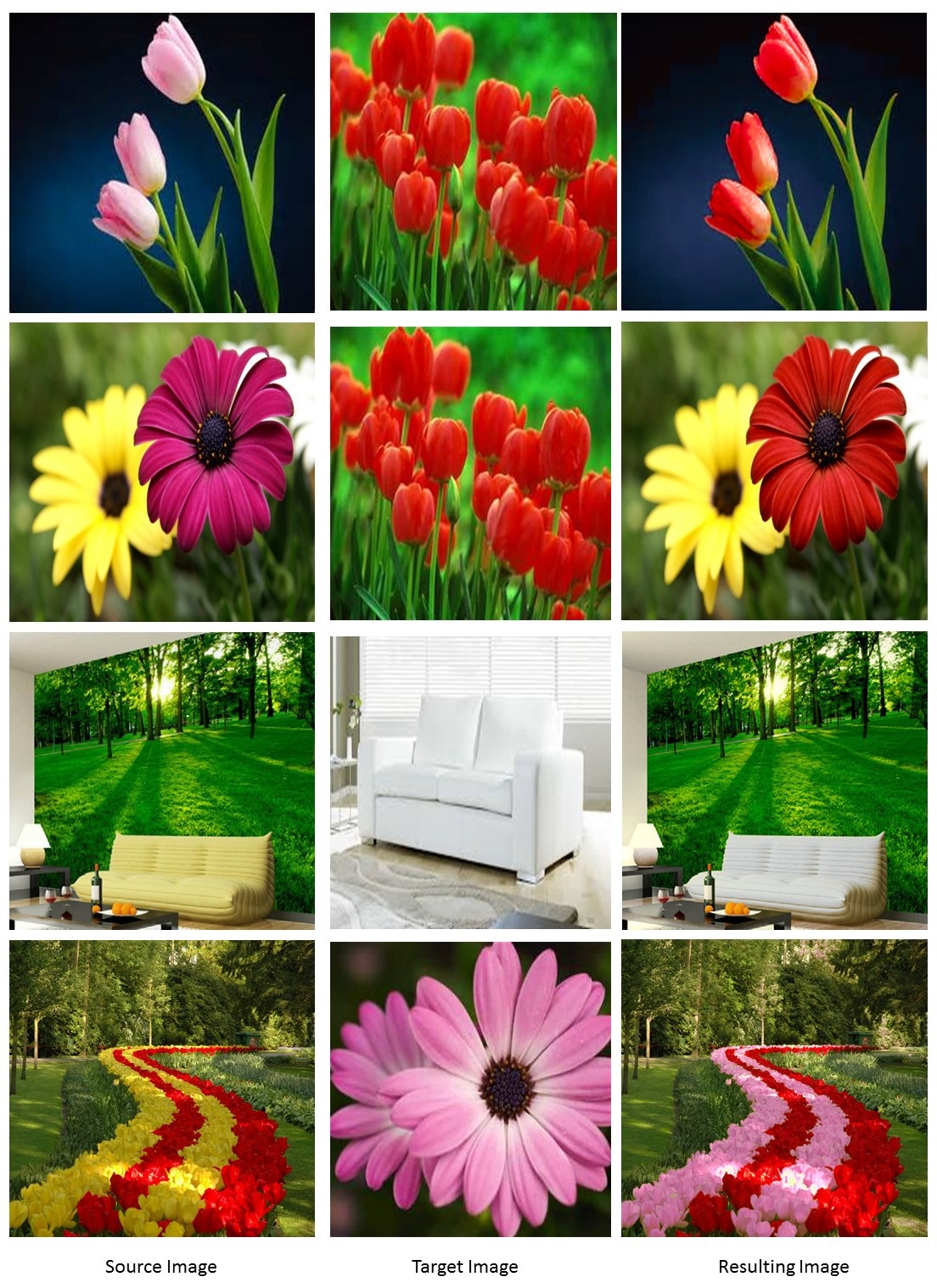}\\
  \caption{Results of one source image and one target image.}
  \label{fig:figure5}
\end{figure*}

Moreover, we compare our results with stokes based techniques depicted in Fig. \ref{fig:figure8}.
Farbman et al. \cite{29} diffuse the local color using the stokes in the first row.
This is a  challenging task due to a high-contrast transitions
between the buildings. Our method propagates the local color efficiently
while preserving the other color details.
Our method produces comparatively the quality result and visual effects as better as by Farbman et al. \cite{29}
While we are having the advantage that our method is a local color transfer between images not stokes
based which clarify it importance.
In a similar fashion, we compare our results with Chen et al. \cite{37} who have also used stokes based technique.
Consequently, our method achieves the same goal with results relatively of same or slightly better quality.\\

We further compare our results with Pouli et al. \cite{32} shown in Fig. \ref{fig:figure9}.
It is clearly seen in first result that while transferring the color to grass they
were not able to preserve the color on tiger which effects the tiger's color.
Moreover, their transferred color on grass is more sharp than its original color in
source image. We transfer the local color on source image more efficiently while preserving the color
of tiger and our technique develops more natural result with better visual effect.
In second result, they have transferred the local color from flower to flower. Here they
could not preserve the color in the carpel part of the flower. Whereas, we were able
to transfer the color while preserving the color in the carpel part of the flower, which
results a more natural result with better visual effects.\\

In Fig. \ref{fig:figure4}, shows some more results to our proposed method using two target images.
They all show the color-transferred results that will reflect the target colors to the source images effectively.
Moreover, the boundary preservation in the resulting image is focused and tackled successfully.
We consider images with different environments like beaches, sceneries and indoor images.
The color has been transferred from sky to sky, flower to flower and shirt to shirt with effective
boundary preservation and quality maintenance.\\

In Fig. \ref{fig:figure5}, shows the result of one source image to one target image, which produces the natural and
color-preserved results in a similar fashion. The color transfer is one-to-one image, though, color has been
transferred merely to those regions which were selected for color transfer.\\

\textbf{Limitation}: One of the limitation of our algorithm is the fact that
we have to give prior instruction to all objects present in the image.
We have to select regions also where we need to maintain the original color besides of selecting
regions where the color is intended to transfer. The results will not be more natural with better
visual effect, if we do not select the regions where the color requires not to be transferred.

\section{Conclusion}
We have presented an algorithm of local color transfer between images based on
the simple statistics and locally linear embedding for edit propagation.
We proposed a window for transferring the local color transfer between the
images. Our suggested technique is very user-friendly and can be applied on commercial scale.
The algorithm is not restricted
to one-to-one image color transfer and can have more than one target images to transfer
the color in different regions in the source image. It is not required by our algorithm
to select the color regions of same styles and same sizes for source and target images. The proposed algorithm
can be applied to a broad range of motives like humans, landscapes,
plants and animals. Overall our method delivers much convincing, faster and user-friendly
algorithm, which generates more natural results with better visual effect.
Comparing with other existing approaches, our method have the same goal but perform better color
blending in the input data. In future, we would like extend this approach in order to process colorization.

\section{References}

\bibliographystyle{model3-num-names}
\bibliography{<your-bib-database>}

\begin{thebibliography}{00}
\bibitem{1} Reinhard E, Ashikhmin M, Gooch B,  Shirley P. Color transfer between images. IEEE Comput Graph Appl 2001;21(5):34-41.
\bibitem{2} Neumann L, Neumann A. Color style transfer techniques using hue lightness and saturation histogram matching. in Computational Aesthetics in Graphics, Visualization and Imaging 2005, p. 111-122, Eurographics Association. 2005.
\bibitem{3} Pitie F, Kokaram A C, Dahyot R. N-dimensional probability density function transfer and its application to color transfer. in IEEE Int Conf on Computer Vision. IEEE Computer Society. 2005; Vol. 2, p. 1434-1439,
\bibitem{4} Pitie F, Kokaram A C, Dahyot R. Automated colour grading using colour distribution transfer. Comput Vis Image Underst 2007;107(1-2):123-137.
\bibitem{5}	Tai Y W, Jia J, Tang C K. Local color transfer via probabilistic segmentation by expectation-maximization. in IEEE Computer Society Conf on Computer Vision and Pattern Recognition 2005;1:p.747-754.
\bibitem{6}Tai Y W, Jia J, Tang C K.  Soft color segmentation and its applications. IEEE Trans Pattern Anal Mach Intell 2007; 29(9): 1520-1537.
\bibitem{7}Kim J H, Shin D K, Moon Y S. Color transfer in images based on separation of chromatic and achromatic colors. in Proc of the 4th Int Conf on Computer Vision Computer Graphics Collaboration Techniques; 2009; p. 285-296
\bibitem{8}Ha H G. Local color transfer using modified color influence map with color category. in IEEE Int Conf on Consumer Electronics-Berlin 2011; p. 194-197.
\bibitem{9}Maslennikova A, Vezhnevets V. Interactive local color transfer between images. in Proc of Graphicon 2007.
\bibitem{10}An X, Pellacini F. User-controllable color transfer. Comput Graph Forum 2010; 29(2): 263-271.
\bibitem{11}Wang B, Yu  Y, Wong TT, Chen C, Xu YQ. Data-driven image color theme enhancement. ACM TOG 2010; 29(6): 146.
\bibitem{12}Welsh T, Ashikhmin M, Mueller K. Transferring color to grayscale images. ACM TOG 2002; 21(3): 277-280.
\bibitem{13}Abadpour A, Kasaei S. A fast and efficient fuzzy color transfer method. in Proc of the Fourth IEEE Int Symp on Signal Processing and Information Technology 2004, p. 491-494.
\bibitem{14}Xiao X, Ma L. Color transfer in correlated color space. in Proc ACM Int  Conf  on Virtual Reality Continuum and Its Applications 2006, p. 305-309.
\bibitem{15}Pitie F, Kokaram  A. The linear Monge-Kantorovitch linear colour mapping for example-based colour transfer. In 4th European Conference on Visual Media Production(IETCVMP) 2007, p. 1-9.
\bibitem{16}Chang Y, Saito S, Nakajima M. Example-based color transformation of image and video using basic color categories. IEEE Trans Image Process 2007; 16(2): 329-336.
\bibitem{17}Chang Y, Saito S, Uchikawa K, Nakajima M. Example-based color stylization of images. ACMTrans  Appl  Percept 2005; 2(3): 322-345.
\bibitem{18}Yang, C.-K., Peng, L.-K.: Automatic mood-transferring between color images. IEEE Comput. Graph. Appl. 28(2), 52-61 (2008)
\bibitem{19}Xiao X, Ma L. Gradient-preserving color transfer. Comput Graph Forum 2009; 28(7): 1879-1886.
\bibitem{20}Wang B, Yu Y, Wong T T, Chen C, Xu Y Q. Data-driven image color theme enhancement. ACM Trans Graph 2010; 29(6): 146.
\bibitem{21}Wang B, Yu Y, Xu Y Q. Example-based image color and tone style enhancement. ACM Trans Graph 2011; 30(4): 64.
\bibitem{22}Cohen O D, Sorkine O, Gal R, Leyvand T, Xu Y Q. Color harmonization. ACM Trans. Graph 2006; 25(3): 624.
\bibitem{23}Shapira L, Shamir A, Cohen O D. Image appearance exploration by model-based navigation. Comput Graph  Forum 2009; 28(2):629-638.
\bibitem{24}Liu  X, Wan L, Qu Y, Wong  T T, Lin S, Leung C S, Heng P A. Intrinsic colorization. ACM Trans  Graph 2008; 27(5): 152.
\bibitem{25}Chia A Y S, Zhuo S, Gupta R K, Tai Y W, Cho S Y, Tan P, Lin S. Semantic colorization with Internet images. ACM Trans Graph 2011; 30(6):1
\bibitem{31}De Silva V, Tenenbaum J B. Sparse multidimensional scaling using landmark points. Technical report Stanford University (2004)
\bibitem{27}Yoo J D, Park M K,Lee K H. Local color transfer between images using dominant colors. Journal of Electronic Imaging 2013;22(3): 033003
\bibitem{28}Criminisi A, Sharp T, Rother C, Perez P.Geodesic image and video editing. ACM Trans Graph 2010; 29(5):134.
\bibitem{29}Farbman Z, Fattal R,Lischinski D. Diffusion maps for edge-aware image editing. ACMTrans Graph 2010; 29(6): 145.
\bibitem{30}Li Y, Adelson E H, Agarwala A. Scribbleboost: Adding classification to edge-aware interpolation of local image and video adjustments. Comput Graph Forum 2008; 27(4):1255-1264.
\bibitem{32}Pouli T, Reinhard E. Progressive color transfer for images of arbitrary dynamic range. Comput Graph 2011; 35: 67–80.
\bibitem{37}Chen X, Zou D, Zhao Q, Tan P. Manifold preserving edit propagation. ACM Trans Graph 2012; 31(6): 132.
\bibitem{38}Musialski P, Cui M, Ye J, Razdan A, Wonka P. A framework for interactive image color editing. Vis Comput 2013; 29:1173-1186.
\bibitem{39}Roweis S T, Saul L K. Nonlinear dimensionality reduction by locally linear embedding. Science 2000; 290(5500): 2323-2326.
\bibitem{40}Saul L K, Roweis S T. Think globally, fit locally: unsupervised learning of low dimensional manifolds. J Mach  Learn Res 2003; 4(2): 119-155.
\bibitem{41}An X, Pellacini F. AppProp: all-pairs appearance-space edit propagation. ACM Trans Graph 2008; 27(3): 40.
\bibitem{42}Lischinski D, Farbman Z, Uyttendaele M, Szeliski R. Interactive local adjustment of tonal values. ACM Trans Graph 2006; 25(3):646.
\end{thebibliography}

\end{document}